\documentclass[12pt]{article}

\usepackage{tabularx}
\usepackage{wrapfig}
\usepackage{titlesec}
\usepackage{mathpazo}
\usepackage{ebgaramond}
\usepackage{charter}
\usepackage{bbold}
\usepackage{booktabs}
\usepackage{array}
\usepackage{multirow}
\usepackage{pifont}
\usepackage{xcolor}

\usepackage{bm}

\definecolor{myblue}{RGB}{13, 71, 161}   
\definecolor{myred}{RGB}{232, 93, 117}   
\usepackage{tikz}
\usetikzlibrary{positioning, arrows.meta}
\usetikzlibrary{calc} 
\usepackage{pdflscape}
\usepackage{adjustbox}
\usepackage{subcaption}
\usepackage{colortbl} 
\usepackage{longtable} 

\definecolor{customgray}{gray}{10.5}  
\newcommand{\mycomment}[1]{}
\titleformat{\section}
{\normalfont\Large\bfseries\color{customgray}} 
{\thesection}{1em}{}

\titleformat{\subsection}
{\normalfont\large\bfseries\color{gray}} 
{\thesubsection}{1em}{}
\titleformat{\subsubsection}
{\color{gray}\normalsize\bfseries} 
{\thesubsubsection}{1em}{}
\titlespacing*{\section}{0pt}{1.5ex plus 1ex minus .2ex}{1.5ex plus .2ex}
\titlespacing*{\subsection}{0pt}{1.25ex plus 1ex minus .2ex}{1.25ex plus .2ex}

\usepackage[isbn=false,url=false]{biblatex}
\usepackage{times}
\usepackage{tabularray}
\usepackage{heuristica}
\usepackage{wrapfig}
\usepackage{multirow}
\usepackage{booktabs}
\usepackage{algorithm}
\usepackage{algpseudocode}
\usepackage{makecell}

\usepackage[hmargin=1cm]{geometry}

\usepackage{tikz}

\usetikzlibrary{backgrounds}
\usetikzlibrary{patterns}
\usetikzlibrary{bayesnet}
\usetikzlibrary{arrows}
\usepackage{color}

\usetikzlibrary{backgrounds}

\usepackage{array}

\usepackage{amsmath,amsfonts}
\usepackage{amssymb} 
\usepackage{mathtools}

\usepackage{listings}
\usepackage[most]{tcolorbox}

\definecolor{usercolor}{RGB}{255, 108, 108} 
\definecolor{assistantcolor}{RGB}{255, 189, 69} 

\newtcolorbox{userbox}{
    colback=usercolor!5!white,
    colframe=usercolor!75!black,
    width=\linewidth-1cm,
    left=0.5cm,
    right=0.5cm,
    boxrule=0.4mm,
    arc=2mm,
    fonttitle=\bfseries,
    title=Prompt,
    breakable
}
\usepackage{comment}
\usepackage{booktabs}

\newtcolorbox{assistantbox}{
    colback=assistantcolor!5!white,
    colframe=assistantcolor!75!black,
    width=\linewidth-1cm,
    right=0.5cm,
    left=0.5cm,
    boxrule=0.4mm,
    arc=2mm,
    fonttitle=\bfseries,
    title=Assistant,
    breakable
}

\lstdefinestyle{pythonstyle}{
    language=Python,
    basicstyle=\ttfamily\small\color{black},           
    keywordstyle=\color{teal}\bfseries,                 
    stringstyle=\color{brown},                          
    commentstyle=\color{gray},                          
    numberstyle=\tiny\color{gray},                      
    frame=single,                                       
    backgroundcolor=\color{gray!5},                     
    breaklines=true,                                    
    showstringspaces=false,                             
    numbers=left,                                       
    xleftmargin=1em,                                    
    rulecolor=\color{black},  
    morecomment=[s][\color{olive!80!black!80}\slshape]{"""}{"""},  
}

\lstdefinestyle{bashstyle}{
    language=bash,
    basicstyle=\ttfamily\small,
    frame=single,
    backgroundcolor=\color{gray!10},
    breaklines=true,
    showstringspaces=false
}

\usepackage{amsmath}
\usepackage{amssymb}
\usepackage{mathtools}
\usepackage{amsthm}
\theoremstyle{plain}
\newtheorem{theorem}{Theorem}[section]

\theoremstyle{definition}

\theoremstyle{remark}

\definecolor{bronze}{RGB}{205, 127, 50}
\definecolor{silver}{RGB}{192, 192, 192}
\definecolor{gold}{RGB}{255, 215, 0}
\usepackage{graphicx}   
\usepackage{amsmath}    
\usepackage{amssymb}    
\usepackage{fancyhdr}   
\usepackage{geometry}   
\usepackage{titlesec}   
\usepackage{xcolor}     
\usepackage{caption}  
\titleformat{\section}{\normalfont\bfseries\Large}{\thesection}{1em}{}
\titlespacing*{\section}{0pt}{*3}{*2}

\newtoggle{final}
\togglefalse{final} 

\newcolumntype{C}{>{\centering\let\newline\\\arraybackslash\hspace{0pt}}m{2cm}}

\newcommand{\OurAgent}{\textrm{Sphere Packing}}

\newcommand{\version}{v1.1}

\newcommand\vartextvisiblespace[1][.5em]{%
  \makebox[#1]{%
    \kern.07em
    \vrule height.3ex
    \hrulefill
    \vrule height.3ex
    \kern.07em
  }%
}

\usepackage{tcolorbox}

\usepackage{url}      
\usepackage{hyperref}   

\definecolor{headerblue}{RGB}{250, 245, 245}
\definecolor{titleblack}{RGB}{80, 0, 0} 
\definecolor{abstractblack}{RGB}{80, 0, 0} 

\newtcolorbox{fullbox}{
colback=headerblue,
colframe=white,
width=\textwidth,
boxrule=0pt,
arc=10pt,
outer arc=10pt,
boxsep=10pt,
left=10pt,
right=10pt,
top=10pt,
bottom=10pt
}

\author
{Firstname Lastname,$^{1}$ 
\\
\normalsize{$^{1}$Huawei Noah’s Ark Lab, London, UK.}\\
\normalsize{$^{2}$Technical University of Darmstadt, Darmstadt, Germany.}\\
\normalsize{$^{3}$University College London, London, UK.}\\
\\
\normalsize{
Correspondence: 
firstname.lastname@huawei.com,} \\
}

\date{}

\begin{document}




\begin{fullbox}
\vspace{0.1em}
\begin{center}
{\Large \textbf{Model-Based and Sample-Efficient AI-Assisted Math Discovery in Sphere Packing}}

\vspace{1em}
\textbf{\textsf{Rasul Tutunov$^{1}$, Alexandre Maraval$^{1}$, Antoine Grosnit$^{1}$, Xihan Li$^{2}$,  Jun Wang$^{2}$, Haitham Bou-Ammar$^{1,2}$}}\\
{\small $^1$ Huawei Noah's Ark} {\small $\ ^2$ AI Centre, Department of Computer Science, UCL}\\

\end{center}

\noindent

\textbf{Abstract:} 
Sphere packing, Hilbert’s eighteenth problem, asks for the densest arrangement of congruent spheres in ${n}$-dimensional Euclidean space. Although relevant to areas such as cryptography, crystallography, and medical imaging, the problem remains unresolved: beyond a few special dimensions, neither optimal packings nor tight upper bounds are known. Even a major breakthrough in dimension  ${n}=8$, later recognised with a Fields Medal, underscores its difficulty.

A leading technique for upper bounds, the three-point method, reduces the problem to solving large, high-precision semidefinite programs (SDPs). Because each candidate SDP may take days to evaluate, standard data-intensive AI approaches are infeasible.
We address this challenge by formulating SDP construction as a sequential decision process, the SDP game, in which a policy assembles SDP formulations from a set of admissible components. Using a sample-efficient model-based framework that combines Bayesian optimisation with Monte Carlo Tree Search, we obtain new state-of-the-art upper bounds in dimensions 4–16, showing that model-based search can advance computational progress in longstanding geometric problems. Together, these results demonstrate that sample-efficient, model-based search can make tangible progress on mathematically rigid, evaluation-limited problems, pointing towards a complementary direction for AI-assisted discovery beyond large-scale LLM-driven exploration.

\end{fullbox}

\thispagestyle{fancy}
Artificial intelligence tools have shown an exceptional ability to solve mathematical problems from a wide spectrum of difficulty levels \cite{AlphaProof, AlphaGeometry2,ren2025deepseekproverv2advancingformalmathematical}. One such tool is AlphaEvolve \cite{novikov2025alphaevolve, FunSearch}, a system that has effectively utilised LLMs to tackle problems that are hard to solve but easy to evaluate (i.e., allow for fast evaluation of candidate solutions). These problems, known in theoretical computer science as NP-complete, involve tasks where determining an exact solution is difficult, yet verifying the quality of any proposed solution is typically straightforward. 

However, NP-complete problems represent only a subset of a larger class known as NP-hard problems, where even verifying candidate solutions can be complex and (frustratingly) slow. One prominent example in such a class is that of sphere packing, which aims to fill the Euclidean space $\mathbb{R}^{\text{n}}$ with spheres of the same or different volumes to maximise their spatial coverage. Despite its seemingly simple and abstract formulation, this problem has seen widespread applications, e.g., in biology~\cite{erickson1973tubular}, chemistry~\cite{MEAGHER201555}, and medicine~\cite{wang1999packing}, to name a few. Finding optimal sphere-packing densities is hard! In fact, it is much harder than local setups like kissing number problems, as sphere packing requires understanding infinitely large configurations, not just a single neighbourhood. The arrangement of spheres in one region must remain consistent with placements arbitrarily far away, creating long-range constraints that do not appear in local problems. As the dimension grows, these global dependencies only become more complex, making density calculations even more difficult.

Interestingly, the roots of the sphere packing problem go back over two millennia, to Euclid’s exploration of space and volume around 300 BC. Yet despite centuries of mathematical progress, the problem remains unsolved in most dimensions. To date, optimal arrangements of identical unit-volume spheres have been discovered only for 
n=2 \cite{fejes1942dichteste}, 3 \cite{hales2017formal}, 8 \cite{viazovska2017sphere}, and 
24 \cite{cohn2017sphere}. The eight-dimensional case, proved by Maryna Viazovska and recognised with the 2022 Fields Medal, highlights the enduring significance of sphere packing in contemporary mathematics.

For dimensions where optimal arrangements are still unknown, researchers have focused on deriving rigorous upper and lower bounds on the maximum packing density, which are available for all dimensions \cite{Spherepacking}. Importantly, improving upper bounds for packing problems has proven to be a powerful strategy for identifying optimal sphere configurations. Notably, Cohn et al.~\cite{Cohn_2003} improved the upper bounds in dimensions $n=8$ and $
n=24$, showing densities close to those of the $E_8$ root lattice and the Leech lattice, respectively. This work provided compelling evidence that these lattices might indeed be optimal and guided subsequent efforts to prove it. Building on this foundation, Viazovska later proved the optimality of the $E_8$ lattice by further refining the upper bound using a distinct method, which was subsequently extended to the Leech lattice case. Thus, improvements in upper bounds not only constrain possible configurations but can also illuminate the path toward exact optimality proofs.

Among other approaches, the three-point method proposed by Cohn et al.~\cite{cohn2022three} attains state-of-the-art results by solving carefully designed semidefinite programming (SDP) problems, where the optimal objective value of those SDPs directly provides an upper bound on the maximum packing density. Each such SDP instance requires two main components: 1) a set of continuous geometric parameters in 2048-bit precision (approximately 617 digits in decimal representation) and 2) a set of specifically crafted constraints formulated in terms of matrix polynomials. The search for these parameters and the corresponding matrix polynomial representations has become a hallmark of recent progress in this area. Different parameterisations yield distinct SDP formulations, each defining a new optimisation landscape whose solutions may yield tighter upper bounds on packing density. 

Given that the discovery of improved sphere packing bounds can be formulated as a search over continuous parameters and polynomial constraint representations, a natural question arises: \emph{Can AI-assisted methods accelerate this search and uncover new state-of-the-art results on such long-standing mathematical problems?} At first glance, recent breakthroughs such as AlphaEvolve \cite{AlphaEvolve} and AlphaGeometry \cite{AlphaGeometry2} suggest so, showcasing the promise of AI in mathematical discovery. However, these systems rely primarily on model-free and evolutionary search strategies, where learning proceeds through extensive trial and error, i.e., evaluating vast numbers of candidates to gradually refine a policy or population. Such approaches thrive in domains where solutions are hard to find but cheap to evaluate. In contrast, our setting lies at the opposite extreme: evaluating a single candidate, by solving its corresponding semidefinite program, can take days. This renders data-hungry, evaluation-intensive methods impractical and calls for more sample-efficient alternatives.
\begin{figure}[t!]
\centering
\includegraphics[trim={0em 0em 0em 7em}, clip=true, width=1\linewidth]{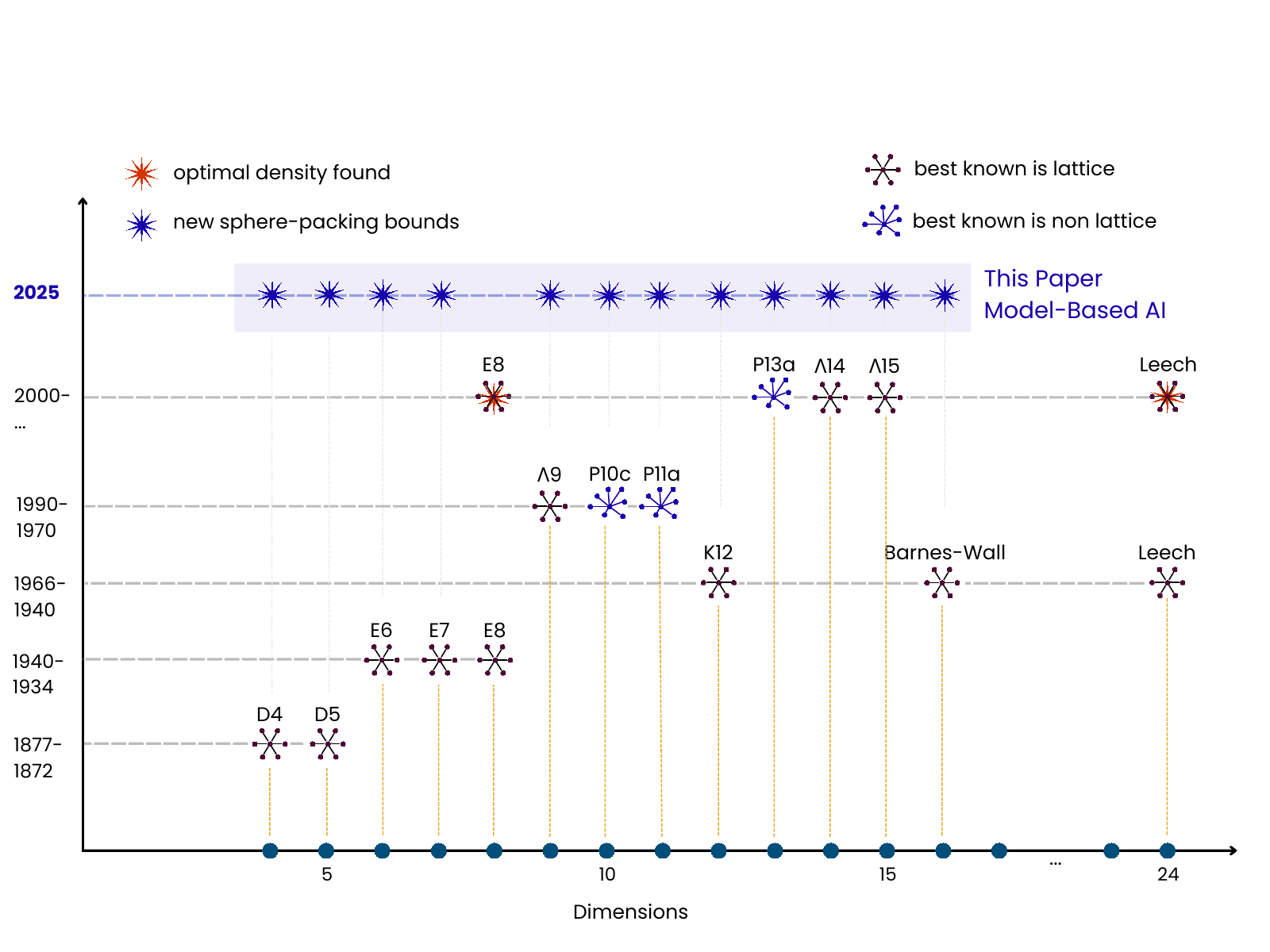}
\caption{\textbf{A Brief History of the best known sphere-packing bounds across dimensions.} This figure highlights the classical milestones in sphere-packing. The range $4 \leq n \leq 16$ has been the most intensely studied, with decades-long progress driven by lattice and non-lattice constructions (see $D_4$, $E_6$, $E_7$, $E_8$, $K_{12}$, Barnes-Wall, Leech, and others). Our model-based AI approach (blue stars, 2025) achieves new state-of-the-art bounds throughout this critical band of dimensions, surpassing all previously known constructions.}
\label{fig:history}
\end{figure}

In this work, we explore an alternative to sample-intensive approaches by introducing a sample-efficient, model-based framework for AI-assisted mathematical discovery in sphere packing. To this end, we formulate the construction of optimal SDP instances as a single-player SDP game. In the first round of this game, the player selects a set of continuous geometric parameters with high-precision values. Given these, in the second round, the player constructs a collection of SDP instances, each defined by an objective function and a set of agent-selected matrix polynomial constraints. The quality of each instance is determined by its optimal objective value, which in turn guides the player’s strategy across both rounds. This formulation defines a challenging game, where the agent needs to determine optimal actions in a mixed search space: continuous in the first round and discrete in the second. 

To solve the SDP-Game efficiently, we develop a sequential, model-based optimisation approach. Continuous parameters in the first round are selected using Bayesian Optimisation (BO), while objective functions and matrix polynomial constraints in the second round are generated through Monte Carlo Tree Search (MCTS). Both rounds leverage progressively refined surrogate models to improve sample efficiency and minimise the number of costly SDP evaluations. Our method, when applied to the classical sphere packing problem, establishes previously unknown upper bounds in Euclidean spaces $\mathbb{R}^n$ for \emph{twelve dimensions}, $n \in [4, 16]$, advancing the best-known limits in this long-standing open problem (Figure \ref{fig:history}). These results highlight the potential of model-based, AI-driven search to uncover mathematical insights that would be computationally prohibitive through manual exploration or model-free approaches.

\section*{Semidefinite Programming (SDP) Games for Sphere Packing}
The sphere packing problem asks for the densest possible arrangement of non-overlapping equal spheres in $\mathbb{R}^{\text{n}}$. Exact optimal configurations are known only for a few dimensions $n=2,3,8$, and $24$, while in all others, only upper and lower bounds are available. To study these cases and improve those bounds, researchers turned to approximation methods. The linear programming (LP) or two-point method, introduced by Cohn and Elkies~\cite{Cohn_2003}, provided powerful upper bounds by leveraging pairwise distance information between sphere centres. However, this approach is incomplete, as it ignores higher-order geometric constraints. To address this limitation, Cohn et al.~\cite{cohn2022three} proposed the three-point method, which extends the LP framework to include interactions among triples of points, leading to significantly tighter bounds that are so far widely considered the tightest known. 

This three-point method showed that determining upper bounds on sphere packing density can be cast as an SDP problem, a powerful generalisation of linear programming. In an SDP, vector variables and coefficients are replaced with matrices, and nonnegativity constraints are extended to positive semidefiniteness conditions, see Methods for an exact mathematical exposition. This formulation allows geometric relationships among multiple points to be encoded through matrix-valued polynomial inequalities, while the objective and constraints are expressed via trace operators over these matrices. Although standard solvers such as SDPA-GMP~\cite{sdpa_solver} can handle such programs reliably, the large number of variables and semidefinite constraints makes them substantially more computationally demanding than linear programs.

\begin{figure}[t!]
\centering
\includegraphics[trim={0em 6em 0em 5em}, clip=true, width=1\linewidth]{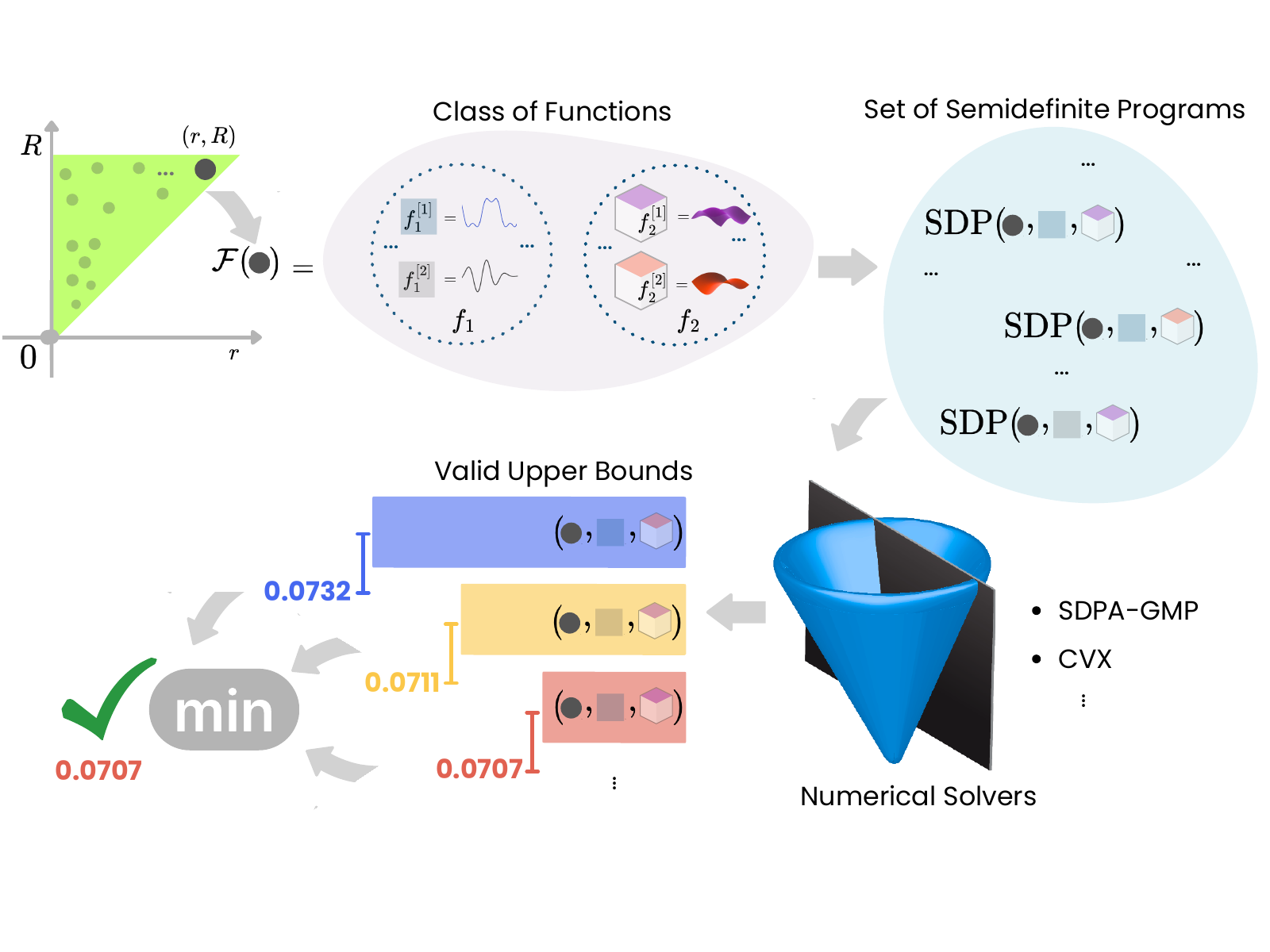}
\caption{\textbf{Three-Point Bound Method for Sphere Packing Problems. } This method establishes upper bounds on the density of arbitrary sphere packings in $\mathbb{R}^{\text{n}}$ by formulating specific semidefinite programming (SDP) problems. The process begins with the choice of two nonnegative geometric parameters $(r,R)$ with $R > r$.  
Given these parameters, one constructs the function class $\mathcal{F}(r,R)$. Each pair $(r, R)$ together with functions $f_1, f_2 \in \mathcal{F}(r, R)$ define a specific semidefinite program $\text{SDP}(r, R, f_1, f_2)$. These SDPs, are then solved using standard SDP solvers, and their optimal objective values provide valid upper bounds on the sphere-packing density.}
\label{fig:three-point_bound}
\end{figure}

Within the three-point framework, the associated SDP is not \emph{uniquely defined}. For any pair of nonnegative geometric parameters  $r, R$ with $r< R$, and any two functions $f_1, f_2$  from a carefully constructed class $\mathcal{F}(r,R)$, one can formulate a semidefinite program $\text{SDP}(r,R,f_1,f_2)$ that yields an upper bound $\Delta_{\text{upper}}(r,R,f_1,f_2)$ on the density of any sphere packing: 
\begin{equation*}
    \textcolor{blue}{\text{Solve}[}\text{SDP}(r,R,f_1,f_2)\textcolor{blue}{]} \rightarrow  \Delta_{\text{upper}}(r,R,f_1,f_2), 
\end{equation*}
where $\textcolor{blue}{\text{Solve}[\cdot]}$ executes any off-the-shelve algorithm to solve $\text{SDP}(r,R,f_1,f_2)$, e.g., SDPA-GMP, or CVX. This overall process is illustrated in Figure \ref{fig:three-point_bound}, which visualises how different choices of $(r,R)$ and $f_1, f_2$ in the class of functions $\mathcal{F}(r,R)$ define distinct SDP instances. Each instance is then solved using standard numerical solvers, and its optimal objective value provides a valid upper bound on the sphere packing density.

We notice that since the SDP formulation is not uniquely defined, one can view the search for the tightest upper bound as the task of \emph{identifying the best possible SDP instance.} In other words, we can recast the problem of improving sphere packing bounds as an optimisation over the space of SDP definitions, spanning $(r,R)$ and $(f_1, f_2) \in \mathcal{F}(r,R)$. This perspective naturally transforms the mathematical challenge into a structured search problem as follows: 
\begin{align}\label{optimal_Sdp_equation}
    \text{SDP}^{\star} \leftarrow \text{Search}_{r,R,(f_1,f_2)\in\mathcal{F}(r,R)}\left[ \textcolor{blue}{\text{Solve}[}\text{SDP}(r,R,f_1,f_2)\textcolor{blue}{]}\right],
\end{align}
where the search process identifies the SDP instance yielding the minimal objective value returned by the off-the-shelf solver. 

Despite its conceptual simplicity, performing this search is highly challenging. The core difficulty arises from the fact that gradients with respect to the optimisation variables 
$(r,R,f_1, f_2)$ are not directly accessible: computing them would require differentiating through the semidefinite solver itself—a process that is both unstable and computationally prohibitive. Moreover, the search space is mixed in nature, combining continuous parameters $(r,R)$ with discrete choices of functions $(f_1, f_2)$, which further complicates the application of standard gradient-based optimisation. One could, in principle, employ model-free or evolutionary strategies similar to those used in AlphaEvolve \cite{AlphaEvolve}, which iteratively improve solutions through repeated evaluations. However, each SDP evaluation can take days, making such trial-and-error approaches, and likewise model-free reinforcement learning, computationally infeasible. 
\paragraph{SDP Games.} We address these challenges by introducing a sample-efficient solution to a hierarchical SDP game, defined as an interaction between an agent and an environment represented by the SDP solver. In this game, the agent proposes candidate SDP formulations, while the environment evaluates each proposal by solving the corresponding SDP and returning a reward that reflects the quality of the resulting upper bound.

Formally, the state of the SDP Game is defined as the history of previously explored configurations, each consisting of the chosen parameters and functions $(r,R, f_1, f_2)$ and their corresponding rewards, i.e., the optimal objective values returned by the SDP solver. The action space is hierarchical. In the first stage, the agent selects the geometric parameters $(r,R)$, which determine the admissible function class $\mathcal{F}(r,R)$. In the second stage, conditioned on this choice, the agent constructs a pair of functions $(f_1,f_2)$, thereby fully specifying an SDP instance to be solved. The solver then evaluates this instance and returns the resulting reward, completing one round of interaction. 

The action space of the SDP Game is therefore divided into two distinct levels. At the first level, selecting $(r,R)$ is relatively intuitive, as it amounts to choosing two continuous values within feasible numerical ranges. The second level, however, is considerably more intricate. Here, the agent must construct a pair of functions $(f_1, f_2)$ from the admissible class $\mathcal{F}(r,R)$. The construction of those functions cannot be arbitrary. In fact, to yield valid upper bounds, they must satisfy a set of admissibility and geometric conditions derived from the structure of the sphere packing problem. 

A central component of our approach lies in how we parameterise those second-stage actions. Rather than selecting these functions directly, which would be intractable due to their infinite-dimensional nature and strict admissibility constraints, we introduce a discrete vocabulary that encodes all valid constructions. Each element of this vocabulary corresponds to a base polynomial building block that is guaranteed to satisfy the analytic and symmetry conditions required by the three-point method. By composing these tokens into finite ``sentences'', the agent can generate any admissible pair $(f_1,f_2)$  while preserving mathematical validity by construction. 

For a given $(r,R)$, the vocabulary consists of base polynomials $P_1$ to $P_n$, along with three auxiliary tokens: an empty-space token ``$\langle$ES$\rangle$'', end-of-sentence token ``$\langle$EOS$\rangle$'' and a product token ``$\langle \ast \rangle$''. The grammar operates as follows: a base polynomial $P_i$ is chosen first, followed by either ``$\langle$ES$\rangle$'', ``$\langle\ast\rangle$'', or ``$\langle$EOS$\rangle$''. For example, the construction of $P^2_3P_2P_5 \ \ P_1P^3_7$ maps to: $P_3 \langle\ast\rangle P_3 \langle\ast\rangle P_2 \langle\ast\rangle P_5\langle \text{ES}\rangle P_1 \langle \ast \rangle P_7\langle \ast \rangle P_7 \langle \ast \rangle P_7 \langle \text{EOS}\rangle.$ This representation transforms a highly constrained functional search into a structured space that can be explored efficiently using machine learning tools such as model-based optimisation algorithms.

At this stage, the reader may wonder why we introduce an explicit ``$\langle \text{ES}\rangle$'' token in our vocabulary. This token separates monomials, giving the agent the structural flexibility needed to define new SDP instances, often beyond those considered in prior mathematical constructions, and thereby broadening the space in which improved bounds may be found. The specific polynomial primitives and admissibility rules underlying our vocabulary are further detailed in the Methods section. 
\begin{figure}[t!]
\centering
\includegraphics[trim={0em 0cm 0em 0em}, clip=true, width=1\linewidth]{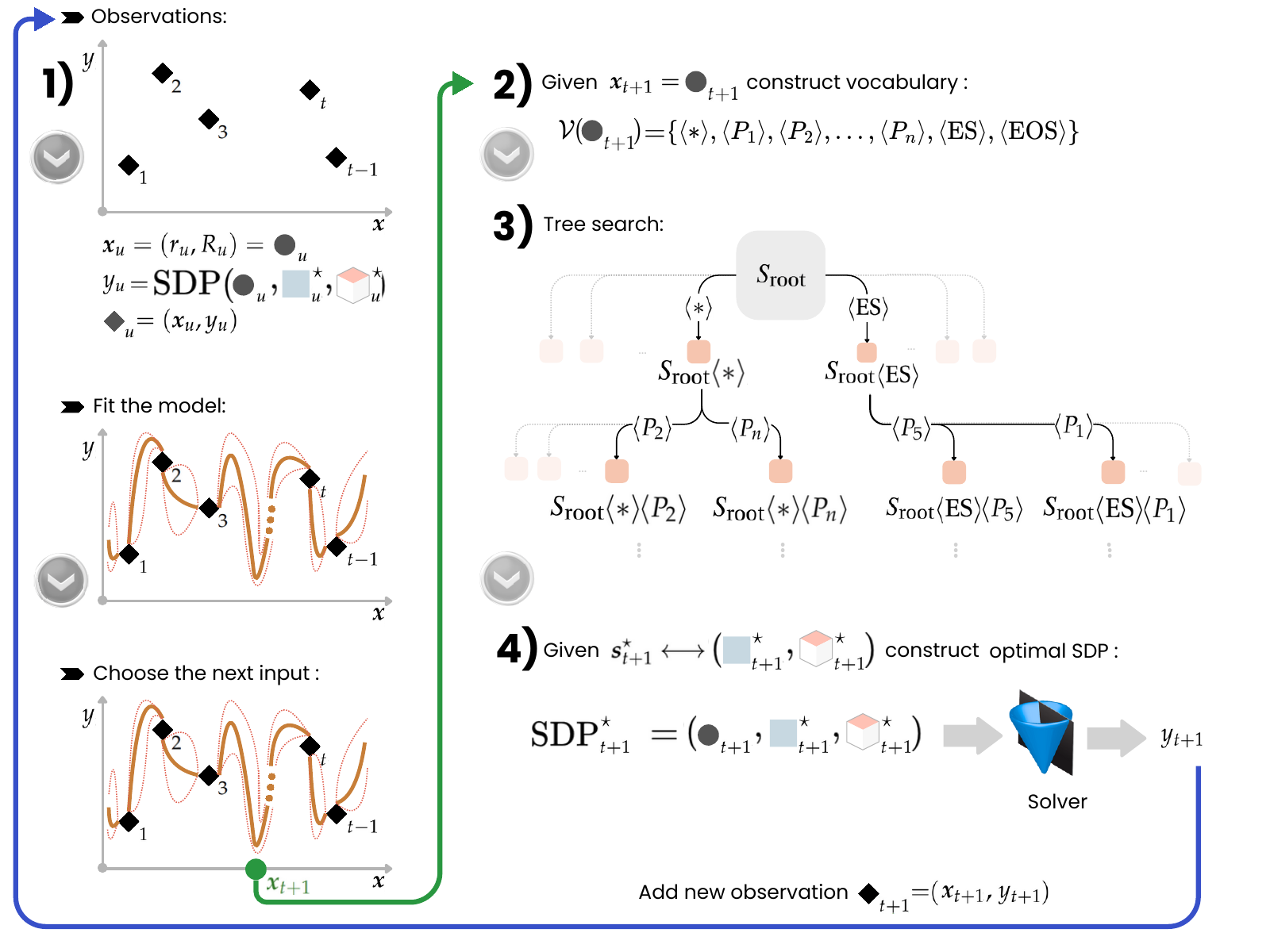}
\caption{\textbf{Overall Framework to Solve SDP Games.} Our approach solves SDP Games through a four-stage, sample-efficient search loop. (1) The agent begins by using Bayesian optimisation to propose geometric parameters $(r,R)$, updating a surrogate model from past evaluations and trading off exploration versus exploitation. (2) Given this choice, it constructs the admissible vocabulary of polynomial building blocks associated with $\mathcal{F}(r,R)$. (3) Monte Carlo Tree Search then explores combinations of these building blocks to assemble candidate SDP formulations that are predicted to yield stronger bounds. (4) The resulting SDP is solved to obtain a certified upper bound, which is added to the dataset and used to guide the next iteration. Through this closed feedback loop, the agent actively selects both the geometric parameters and the polynomial structure of the certificate, enabling it to discover progressively tighter sphere-packing bounds.}
\label{fig:SDP_game_overviewh}
\end{figure}

\section*{Model-Based Solvers for Sphere Packing Bounds}
To efficiently play SDP games, we propose a model-based approach (Figure \ref{fig:SDP_game_overviewh}) that naturally mirrors the game's hierarchical structure. Rather than relying on repeated trial-and-error evaluations (e.g., AlphaEvolve \cite{AlphaEvolve}), which are infeasible due to the cost of solving a single SDP, we employ learned surrogate models that guide the search toward promising regions of the action space.

\paragraph{Bayesian Optimisation for Geometric Parameters.} In the first stage, we optimise $(r,R)$ using Bayesian optimisation (BO). BO is particularly well-suited for settings in which each evaluation is expensive: it builds a probabilistic surrogate of the reward function, namely, the SDP’s optimal objective value returned by the solver, and uses this surrogate to identify the next promising $(r,R)$ to evaluate. This allows the agent to find high-performing parameters after only a small number of costly SDP calls. The surrogate’s uncertainty estimates are crucial here, as they help the agent decide where an additional expensive evaluation is most valuable, balancing exploration of new regions with exploitation of known promising ones.

\paragraph{MCTS for Function Construction.} Conditioned on the chosen $(r,R)$ that defines the class of admissible functions $\mathcal{F}(r,R)$, the second stage constructs the optimal pair $(f_1,f_2)$ using Monte Carlo Tree Search (MCTS), which played a central role in systems such as AlphaGo \cite{silver2016mastering}. A tree node represents an environmental state, expressed as a sequence of monomials. Each edge corresponds to a token that induces a transition to a subsequent state. These transitions respect the grammar rules described above, and a new state is formed by concatenating the preceding state with the chosen token. MCTS iteratively builds a search tree through four phases: selection, expansion, simulation, and backpropagation, learning to balance exploration of new actions with exploitation of those known to yield high rewards. During selection, the algorithm follows a tree policy to choose the most promising path among previously explored nodes. In the expansion phase, a new child node is added by sampling a feasible next token. Simulation then produces a provisional estimate of the final reward by completing the function using the policy model. Finally, backpropagation updates all nodes along the traversed path with the simulated reward, allowing future decisions to incorporate the information gained from this rollout.


\paragraph{Iterative Model-Based Loop.} BO and MCTS interact in an iterative loop: BO proposes a new $(r,R)$; MCTS constructs the best $(f_1,f_2)$ for that choice; solving the resulting SDP yields a new data point; and BO updates its surrogate accordingly; see Figure \ref{fig:SDP_game_overviewh}. This process allows the agent to progressively refine its understanding of where high-quality SDP formulations lie while using only a limited number of expensive evaluations.


Taken together, these components form a unified model-based search procedure for the SDP Game. BO identifies promising regions, while MCTS explores the discrete space of admissible functions. By combining these two levels of reasoning, the agent concentrates computational effort on the most promising SDP formulations, achieving a degree of sample efficiency that would be impossible with model-free or evolutionary alternatives.

\section*{New Upper Bounds for Sphere Packing}
We evaluate our model-based framework on the classical problem of upper bounds for
sphere packing across a wide range of dimensions. The goal of our experiments is twofold:
(1) to determine whether the SDP Game agent can autonomously discover SDP formulations
that improve on the best known bounds in the literature, and (2) to analyse the mathematical
structure of the resulting constructions, assessing both their consistency with prior work
and the extent to which genuinely new terms emerge. Because each SDP evaluation may take days, the experimental setting is highly resource-constrained; nonetheless,
our method identifies new record-breaking bounds in multiple dimensions and reveals
previously unseen polynomial structures that contribute to these improvements.

The results in Table \ref{tab:monomial-analysis} show that our model-based agent consistently improves the best known upper bounds for sphere packing in 
dimensions $n\in[4,7]\cup[9,16]$. These improvements exceed those given by the classical linear-programming bounds and also those from the standard three-point method of \cite{cohn2022three}. \emph{To the best of our knowledge, these represent the strongest upper bounds obtained to date in all corresponding dimensions.}

Beyond numerical improvements, our approach also reveals new
mathematical monomials to consider. The final column reports the percentage of 
monomials in the agent’s construction that do not appear in prior state-of-the-art formulations. Across all dimensions, $80$–$85\%$ of the monomials are newly discovered, indicating that the agent routinely explores regions of the function space that have not been previously examined. At the same time, it reliably recovers the core monomials 
identified by human experts, confirming that the search procedure is both principled and structurally meaningful.

\begin{table}[t]
\centering
\caption{\textbf{Comparison of new upper bounds and monomial structure.} Our method yields new upper bounds on the optimal sphere-packing density in dimensions $n\in[4,7] \cup [9,16]$, surpassing all previously known results, including the classical linear-programming bounds and the optimised three-point method constructions of \cite{cohn2022three}. In every dimension, the AI-based approach matches or improves the best available bound. The final column reports the proportion of newly discovered monomials—terms absent from prior human-designed constructions. Our solution explores broader monomial setups, highlighting the ability of automated search to uncover richer polynomial structures than previously known.}
\vspace{0.5em}
\begin{tabular}{cccc c}
\toprule
\textbf{Dim.\ $n$} & \textbf{LP \ Bound} & \textbf{Cohn et al.} 
& \textbf{AI-Based Bound (Our Method)} & \textbf{New Poly. (\%)} \\
\midrule
4  & 0.64771 & 0.63610733 & 0.63610\textcolor{blue}{\textbf{277}} & 83.3 \\
5  & 0.52498 & 0.51264513 & 0.512\textcolor{blue}{\textbf{2645}} & 85.1 \\
6 & 0.41767 & 0.41030328 & 0.410\textcolor{blue}{\textbf{2905}} & 84.7 \\
7 & 0.32746 & 0.32114711 & 0.3211\textcolor{blue}{\textbf{1475}} & 85.1 \\
9 & 0.19456 & 0.19112042 & 0.19\textcolor{blue}{\textbf{099063}} & 83.7 \\
10 & 0.14759 & 0.14341009 & 0.1434\textcolor{blue}{\textbf{0271}} & 83.3 \\
11 & 0.11169 & 0.10672529 & 0.10672\textcolor{blue}{\textbf{201}} & 80.8 \\
12 & 0.08378 & 0.07971177 & 0.0797\textcolor{blue}{\textbf{066}} & 80.0 \\
13 & 0.06248 & 0.06016444 & 0.0601\textcolor{blue}{\textbf{2103}} & 84.4 \\
14 & 0.04637 & 0.04506122 & 0.0450\textcolor{blue}{\textbf{1031}} & 82.5 \\
15 & 0.03425 & 0.03375644 & 0.0337\textcolor{blue}{\textbf{2585}} & 81.6 \\
16 & 0.0252 & 0.02499441 & 0.0249\textcolor{blue}{\textbf{2121}} & 82.9 \\
\bottomrule
\end{tabular}
\label{tab:monomial-analysis}
\end{table}

\paragraph{Structure of the Newly Discovered Polynomials.} To better understand what our agent actually discovered and why these discoveries enabled it to improve known bounds, we analysed the structural properties of all newly identified auxiliary polynomials across $\text{n}$. 
\begin{wrapfigure}{r}{0.5\textwidth}
  \begin{center}
\includegraphics[trim={0em 3em 0em 1em}, clip=true, width=0.48\textwidth]{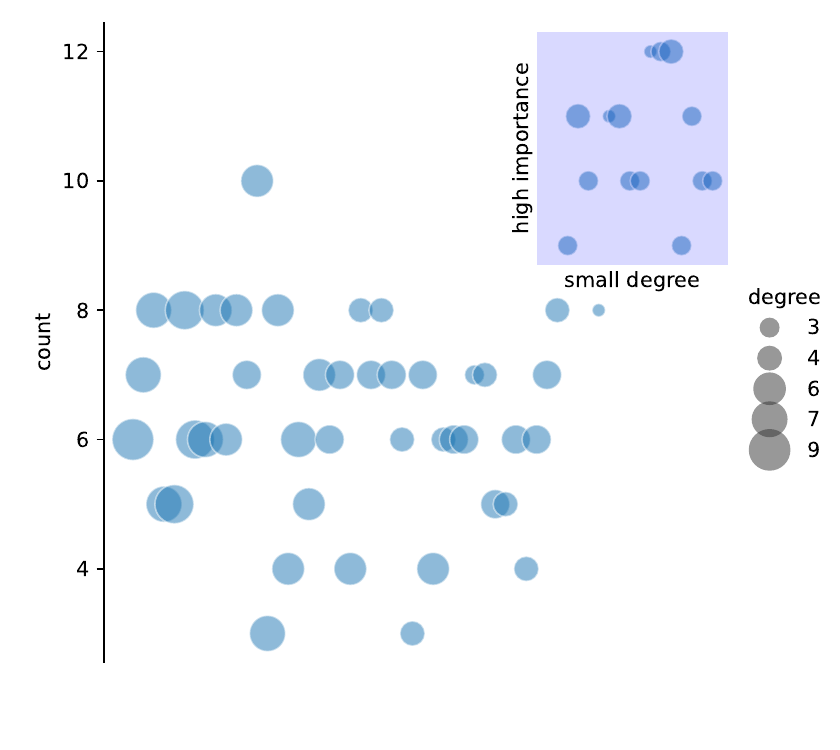}
  \end{center}
  \caption{Scatter plot showing the degrees of polynomials discovered across all dimensions. Each point is a polynomial, with marker size proportional to its degree. We observe that the polynomials that appear most frequently, those with the highest counts (around 12), tend to have low degrees.}
  \label{Fig:Scatter}
\end{wrapfigure}
Although the agent was free to explore a large space of candidate functions, our diagnostic revealed an interesting pattern: the polynomials that appeared most frequently across high-performing SDP solutions were those of low degree. The scatter plot in Figure \ref{Fig:Scatter} illustrates this effect clearly: newly discovered polynomials with the highest occurrence counts cluster at the smallest degrees. 

This observation is mathematically meaningful in the context of the semidefinite program underlying the sphere-packing bound. Low-degree polynomials correspond to variables residing in higher-dimensional SDP blocks, providing the optimisation procedure with significantly more freedom to alter the feasible region. Intuitively, these directions in function space are ``more expressive'' within the constraints of the SDP, and the agent repeatedly returned to them because they offered the most leverage for tightening the packing bound. In other words, while the agent did discover genuinely new polynomial structures beyond those previously explored by human researchers, the analysis suggests that its improvements arose not from isolated exotic functions but from systematically exploiting flexible, low-degree components of the SDP formulation. This provides a new lens on why its search strategy succeeded: it implicitly learned where the SDP is most sensitive and steered its exploration toward those high-impact regions of the functional space. With that said, the analysis also reveals a complementary behaviour: beyond exploiting these highly expressive, low-degree components, the agent occasionally incorporated higher-degree polynomials, even though they appeared less frequently. These rarer functions contributed additional fine-grained adjustments to the SDP, enabling the agent to refine the bound further. 

\begin{figure}[t]
    \centering
    \begin{minipage}[b]{\textwidth}
        \centering
        \begin{subfigure}[b]{0.45\textwidth}
            \centering
            \includegraphics[trim={1em 0em 0em 5em}, clip=true, width=\textwidth]{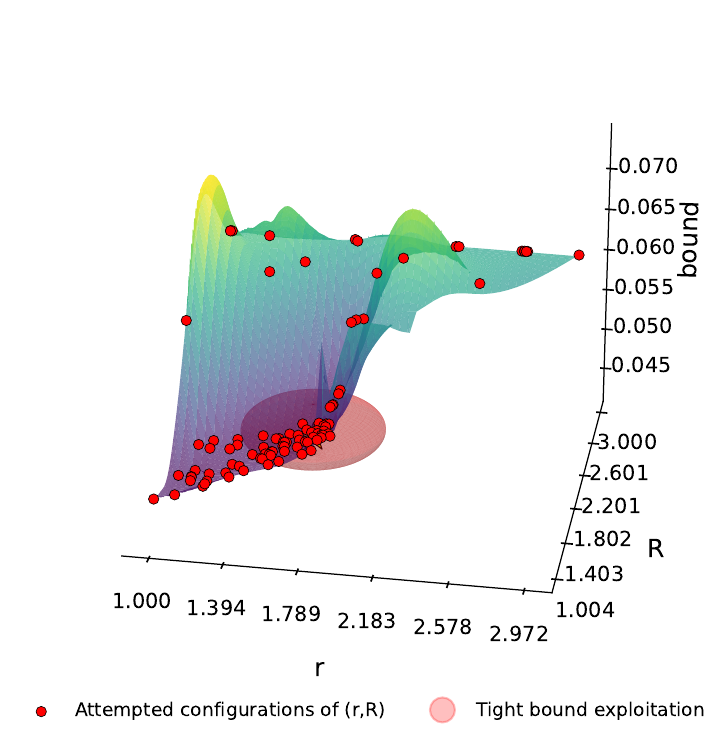}
        \end{subfigure}
        \begin{subfigure}[b]{0.45\textwidth}
            \centering
            \includegraphics[trim={1em 0em 0em 5em}, clip=true,width=\textwidth]{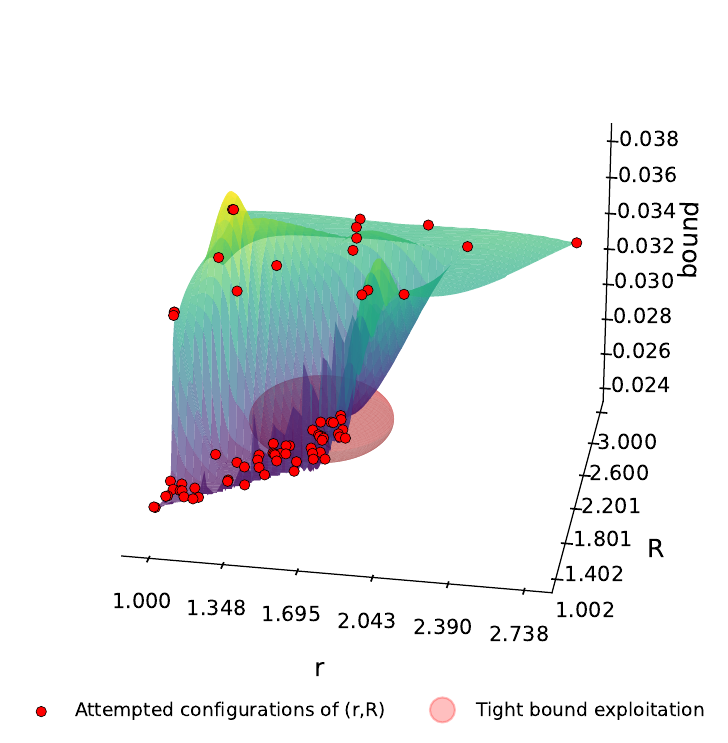}
        \end{subfigure}
    \end{minipage}
    \caption{\textbf{Explored Regions of $(r,R)$ configurations in dimensions 14 (left) and 16 (right).} Our system begins by broadly sampling candidate $(r,R)$ pairs (dark red points), before progressively shifting toward exploitation in regions predicted to yield improved upper bounds (light red region). The shaded red region denotes the neighbourhood of $(r,R)$ values where the tightest bounds have been achieved. Unlike prior mathematical approaches, which varied only a single parameter and therefore explored a restricted slice of the landscape, our method systematically searches across the space and uncovers configurations that were previously unexamined.}
    \label{fig:hebo_r_optim_expensive}
\end{figure}
\paragraph{Expanding the Search Beyond the Classical $r=1$ Regime.} Apart from searching over polynomial structure, our method also searches over the geometric parameters $(r,R)$ that define the admissibility set $\mathcal{F}(r,R)$. This allows us to test whether the agent identifies regions of the parameter space beyond those traditionally considered critical in the sphere-packing literature, where mathematicians typically fix $r=1$ and vary only $R$ \cite{cohn2022three}. By releasing this constraint and optimising jointly over both parameters, the agent can explore a much richer landscape. As shown in Figure \ref{fig:hebo_r_optim_expensive}, the optimiser begins by sampling broadly across $(r,R)$ before concentrating on regions associated with tighter bounds. These high-performing regions, including those in the shaded neighbourhood, lie outside the one-dimensional slices historically examined, demonstrating that our approach uncovers previously unexplored structural regimes that meaningfully improve the resulting bounds.
\paragraph{The $n=8$ Case Study.} In the results presented above, we did not include the $n=8$ case because the optimal sphere-packing bound in eight dimensions is already known \cite{viazovska2017sphere}. This was a landmark result that contributed to Maryna Viazovska receiving the Fields Medal in 2022. To evaluate the capability of our method in a setting where the true optimum is established, we conducted a dedicated case study for 
$n=8$. Our goal was to assess whether the method can recover bounds that approach this optimal value without prior knowledge of modular forms or the special structure underlying Viazovska’s solution, relying solely on the search procedure itself.

Viazovska’s solution to the sphere-packing problem in eight dimensions is based on constructing a special radial Schwartz function: $\mathbb{R}^{8} \rightarrow \mathbb{R}$ with a very specific sign pattern. This function satisfies the following properties:
\begin{enumerate}
    \item \textbf{Non-positivity beyond $||\textbf{x}|| \geq \sqrt{2}$.} The targeted function, $g$ should behave as: 
    \begin{equation*}
        g(\textbf{x}) \leq 0, \ \ \text{for} \ \ ||\textbf{x}|| \geq \sqrt{2}.
    \end{equation*}
    \item \textbf{Non-negativity of the Fourier transform.} The Fourier transform $\hat{g}(\textbf{x})$ of $g(\textbf{x})$ satisfies: 
    \begin{equation*}
        \hat{g}(\textbf{x}) \geq 0, \ \ \text{for all $\textbf{x} \in \mathbb{R}^{8}$}. 
    \end{equation*}
            \item \textbf{Normalisation at the origin.} At the origin both $g(\textbf{x})$ and its Foruier transform $\hat{g}(\textbf{x})$ normalise, such that: 
            \begin{equation*}
                g(\textbf{0}) = \hat {g}(\textbf{0}) = 1.
            \end{equation*}
\end{enumerate}
Finally, both $g(\textbf{x})$ and its Fourier transform are required not to vanish everywhere except $||\textbf{x}||^2 \in 2\mathbb{Z}_{>0}$. 
Intuitively, this means that the function is not permitted to cross zero at arbitrary radii: its zeroes are “locked’’ to the geometric structure of $E_8$. 

In our case study for $n=8$, our goal is not to reproduce Viazovska’s construction or its modular-form structure. Instead, we ask whether the search procedure can automatically discover functions that approximately satisfy this same pattern of sign constraints and Fourier behaviour, using no prior knowledge about $E_8$ or the analytic structure above. Moreover, we test whether the method can produce sphere-packing bounds that improve upon the classical linear-programming bounds and move numerically closer to the known optimal value.
Success on both fronts would demonstrate that the approach can naturally converge toward the optimal packing density in eight dimensions purely from the search dynamics. 

\begin{figure}[t]
    \centering
    \includegraphics[width=\textwidth]{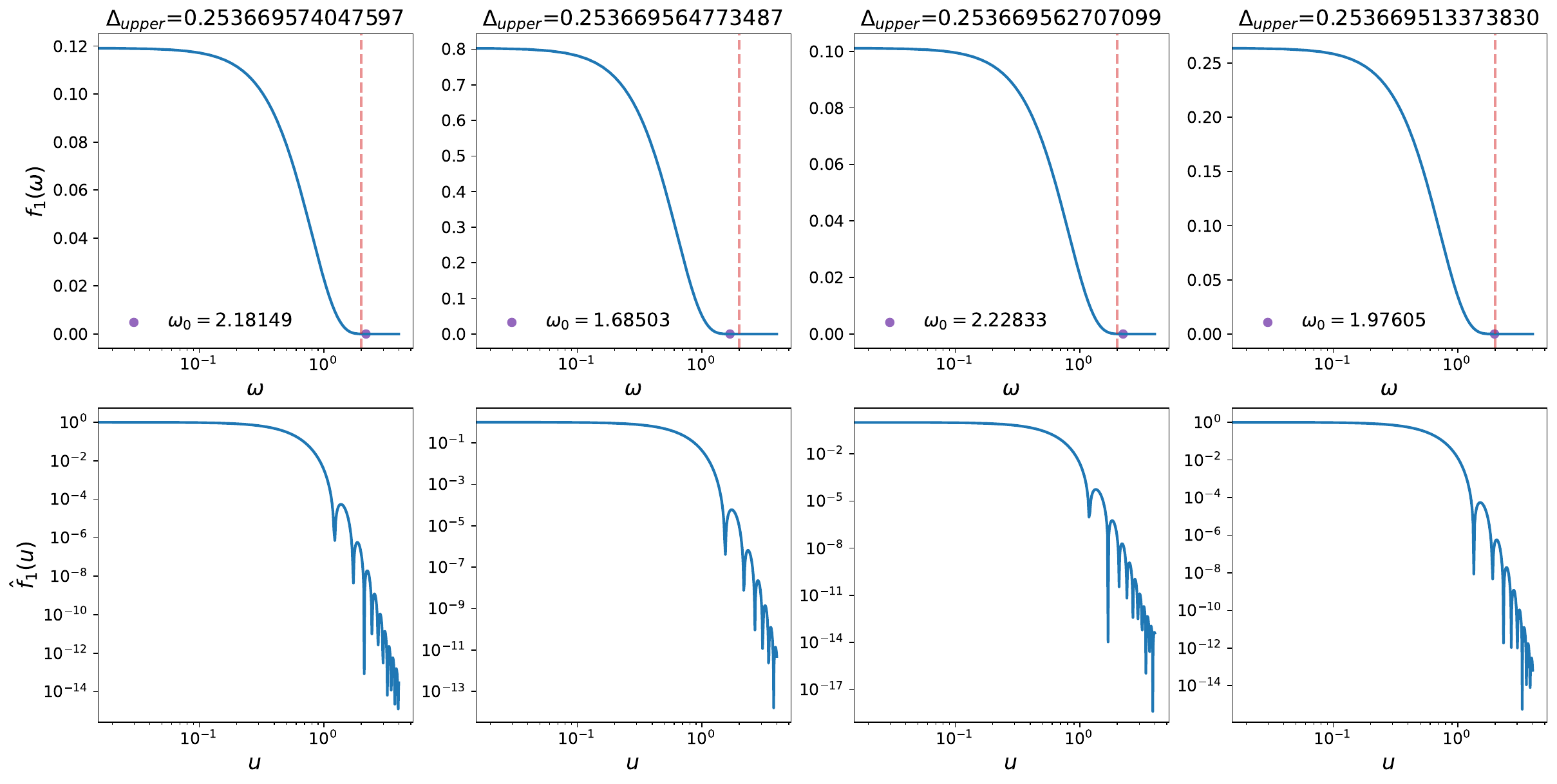}
    \caption{\textbf{Properties of the Discovered Functions.} Plot of the discovered functions (first row) and their Fourier transforms (second row). From the first row, we notice that our algorithm finds functions whose roots approach $\sqrt{2}$. From the second, we notice that the Fourier transform of each of those functions is normalised, $\hat{f}_1(\textbf{0})=1$, and is non-negative.}
    \label{fig:WhatAboutTheFuncs}
\end{figure}
Figure \ref{fig:WhatAboutTheFuncs} demonstrates the properties of the discovered function (first row) in $n=8$ along with its Fourier transform (second row). We order each column by the quality of the resulting upper bound such that the first column on the left represents a worse bound compared to the last column. Focusing on the first row, we notice that as the upper bound improves, the non-positivity region approaches $||\textbf{x}|| \geq \sqrt{2}$, which is consistent with the first condition above. Moreover, examining the second row reveals that the Fourier transform is properly normalised ($\hat{f}_1(\textbf{0})=1$) and is consistently non-negative, i.e., $\hat{f}_1(\textbf{x}) \geq 0$. Interestingly, we further notice that the remaining condition of the normalised function ($f_1(\textbf{0})=1$) is not yet met in our evaluation. This is because the zero of the function is not equal to $\sqrt{2}$, albeit close enough. Theoretically speaking, when the root becomes $\sqrt{2}$, then $f_1(\textbf{0}) \geq 1$. As we currently observe that $f_1(\textbf{0}) \leq 1$, we conjecture that our algorithm will converge closer to a $\sqrt{2}$ root with more iterations, and thus increase $f_1(\textbf{0})$ towards unity. 

In terms of the quality of the discovered bounds in $n=8$, we find that, indeed, our approach produces state-of-the-art results that are closer to the optimal density of $0.2536695079$, where: 1) LP-bound - 0.253670, 2) three-point method \cite{cohn2022three} - 0.2536699179, and 3) ours (SDP Games) - 0.253669\textcolor{blue}{5134}.  

\section*{Discussion and Future Work} 
Artificial intelligence and pure mathematics are increasingly converging, with recent systems such as AlphaGeometry \cite{AlphaGeometry2}, AlphaEvolve \cite{AlphaEvolve}, and DeepSeek-Prover \cite{xin2024deepseek} demonstrating that machine learning can assist in problems traditionally dominated by human insight. However, these successes rely heavily on sample-intensive search, i.e., evaluating large populations of candidate solutions, or performing millions of trial-and-error steps guided by large language models. Such approaches are effective when evaluation is cheap, but become infeasible in domains where every evaluation carries a significant computational cost. Sphere packing represents precisely this opposite regime: each SDP evaluation may require days, and the analytic constraints governing admissible solutions are rigid, delicate, and highly structured.

Our work illustrates that progress in such settings requires a paradigm shift toward sample-efficient mathematical discovery, where models reason strategically about which candidates to test rather than relying on brute-force exploration. By reframing the construction of three-point bounds as a sequential decision process—the SDP Game—and using model-based optimisation, we show that meaningful advances can be achieved within tight evaluation budgets. This paper, therefore, presents a complementary direction for AI-mathematics research: moving beyond sample-intensive trial-and-error toward principled, sample-efficient search mechanisms tailored to structured analytic problems.

Our method achieves new state-of-the-art upper bounds for sphere packing in twelve dimensions and, in doing so, uncovers mathematical structure beyond previously explored regimes. As seen in Table \ref{tab:monomial-analysis}, the agent routinely discovers 80–85\% new monomial terms compared with prior constructions. yet consistently recovers the core, human-designed components essential to the three-point method. The structural analysis (Figure \ref{Fig:Scatter}) reveals that many of these newly identified polynomials are of low degree. Interestingly, those are precisely the components that exert maximal influence in the SDP formulation, while higher-degree terms appear selectively to refine fine-grained constraints. These findings suggest that the agent is not merely exploring blindly but is implicitly learning which regions of the function space provide the greatest leverage for tightening bounds.

The exploration of geometric parameters $(r,R)$ further highlights this effect. Classical treatments fix $r=1$ and optimise only over $R$, effectively restricting the search to a one-dimensional slice. In contrast, our agent discovers high-performing configurations across the broader two-dimensional landscape (Figure \ref{fig:hebo_r_optim_expensive}), identifying regions that had not been considered in prior mathematical work. This behaviour not only leads to improved upper bounds but also points toward new geometric regimes worthy of further analytic attention.

Our case study in dimension $n=8$, where the optimal packing density is exactly known through Viazovska’s remarkable work, provides an additional validation of the approach. The method can recover functions that increasingly resemble the sign constraints and Fourier-space behaviour underlying the magic function, despite not knowing modular forms or the special algebraic properties of $E_8$. Moreover, the resulting SDP constructions produce bounds that improve upon classical linear-programming results and numerically approach the optimal value. These findings suggest that model-based search can autonomously approximate deep analytic features when guided solely by the geometry of the optimisation landscape.

At the same time, several limitations remain. The cost of solving large SDPs fundamentally restricts the number of evaluations available, and scaling to higher-dimensional settings remains computationally demanding. While our surrogate models provide strong guidance under limited data, they are still approximations of an immensely complex landscape. Furthermore, SDP solvers themselves form a computational bottleneck; improving solver efficiency or integrating differentiable surrogates remains an open challenge.

Looking ahead, we identify several promising directions. First, extending this framework to higher-point bounds could reveal new structural patterns and potentially push known limits in dimensions where little progress has been made. Second, the algebraic patterns discovered by the agent, particularly low-degree structural motifs that appear across multiple dimensions, may offer clues for new analytic conjectures. Third, hybrid approaches that combine symbolic constraints suggested by mathematicians (e.g., vanishing radii or symmetry conditions) with learned exploration over polynomial structures could yield more interpretable certificates. Fourth, developing fast surrogate solvers or learned approximations to the SDP evaluation step could enable larger-scale searches and richer experimentation. Finally, the broader methodology may generalise to other geometric and analytic problems such as energy minimisation, coding theory, or optimal transport inequalities, where structured certificates play a central role.

Our results highlight the potential of sample-efficient, model-based AI to contribute to pure mathematics. Rather than replacing human reasoning or relying on massive computational sweeps, this work illustrates how strategic search with principled inductive biases can uncover new mathematical insights in settings that are too rigid, delicate, or expensive for brute-force exploration. We believe that this shift toward intelligent sampling, structured search, and deeply constrained reasoning will be essential as AI continues to engage with some of the most challenging and beautiful problems in mathematics.


\section*{Methods}
\subsection*{The Three-Point Method for Sphere Packing Bounds}
The three-point method is an analytic framework that currently achieves the strongest known lower bounds on sphere packings. The core of the method stems from the following result proved in \cite{cohn2022three}:
\begin{equation*}
     \Delta_{\text{upper}}(r,R, f_1,f_2) \propto \left(f_1(\bm{0}) + f_2(\bm{0}, \bm{0})\right)r^{\text{n}}.
\end{equation*}
The above result states that we can upper-bound sphere packings in proportion to the parameter $r$, and the two functions $f_1$ and $f_2$. The behaviour of these functions is governed by the parameters $r$ and $R$, through a set of mathematical requirements that can be divided into \textit{individual} conditions and \textit{joint} conditions:
\begin{align*}
    &\textbf{Conditions on } \ f_1: \ \  \begin{cases}
        f_1(\bm{a}) \text{ is continuous and integrable on }\mathbb{R}^{\text{n}}, \\
        \text{ Fourier transform $\hat{f}_1(\bm{b})$ has: } \hat{f}_1(\bm{0}) = 1, \ \hat{f}_1(\bm{b}) \ge 0 \ \forall \bm{b}\in\mathbb{R}^{\text{n}}, \\
        f_1(\bm{a}) \le 0 \text{  for all} \ \bm{a}: ||\bm{a}|| \ge R. 
    \end{cases}
\end{align*}
\begin{align*}
    &\textbf{Conditions on } \ f_2: \ \  \begin{cases}
        f_2(\bm{a},\bm{b}) \text{ is positive definite as  a kernel}, \\
        f_2(\bm{a},\bm{b}) \le 0 \text{ for all $(\bm{a}, \bm{b})$}: \  r\le ||\bm{b}||,||\bm{b}|| \le R, \ \& \  ||\bm{a} - \bm{b}||\ge r. 
    \end{cases} \\
      &\textbf{Joint conditions on } \ f_1 \& f_2: \ \  f_1(\bm{a}) + 2f_2(\bm{0}, \bm{a}) + f_2(\bm{a}, \bm{a}) \le 0   \text{  for all $\bm{a}$ :} \ r\le ||\bm{a}|| \le R
\end{align*}
  
\begin{figure}[t!]
\centering
\includegraphics[trim={0em 4.5cm 0em 5.5cm}, clip=true, width=1\linewidth]{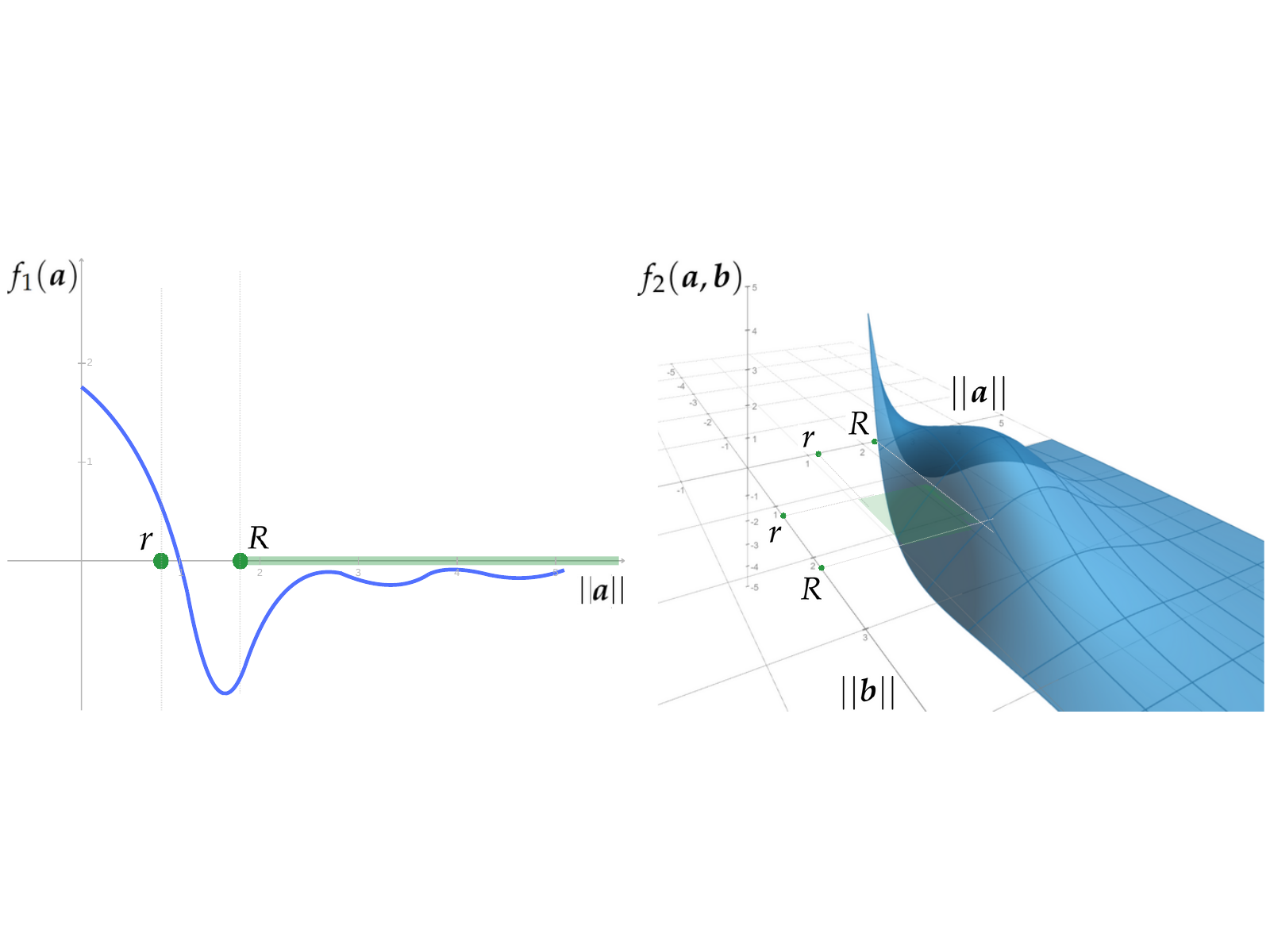}
\caption{Example of functions $f_1$ and $f_2$ satisfying all required mathematical conditions. The function $f_1(\bm{a})$ is radial —depending only on the norm  $||\bm{a}||$)— and has a single sign change. Function $f_2(\bm{a},\bm{b})$ is non-negative on the rectangular region $(\bm{a},\bm{b}): (||\bm{a}||, ||\bm{b}||) \in [r,R]^2$. Both functions decay exponentially to zero as the norms of their arguments increase.}
\label{fig:func_f1andf2}
\end{figure}

The first group includes general properties of functions, such as continuity, integrability, and positive definiteness\footnote{A function $\hat{f}_1(\bm{a}) = \int_{\mathbb{R}^{\text{n}}}f_1(\bm{b})e^{-2i\pi \bm{b}^{\mathsf{T}}\bm{a}}\text{d}\bm{b}$ is called Fourier transform of function $f_1$.} The second group of conditions constrains the behaviour of the functions $f_1$ and $f_2$ in specific regions of their domains determined by the parameters $r$ and $R$. Figure \ref{fig:func_f1andf2} illustrates examples of functions $f_1$ and $f_2$ that satisfy the above conditions. Importantly, each admissible pair of functions $f_1$ and $f_2$ can be represented as a sequence of monomials constructed from the vocabulary of tokens:
\begin{align*}
    \mathcal{V}(r,R) = \{\langle\ast\rangle, \langle P_1\rangle,\langle P_2\rangle,\langle P_3\rangle,\langle P_4\rangle,\langle P_5\rangle,\langle P_6\rangle,\langle P_7\rangle, \langle\text{ES}\rangle, \langle\text{EOS}\rangle \}
\end{align*}
where the symbols $P_i$ represent polynomials of at most three variables $(h,v,w)$ symmetric with respect to the first and second arguments \footnote{A polynomial $q(h,v,w)$ is symmetric with respect to the first and second arguments implies $q(h,v,w) = q(v,h,w)$.}, and take the following explicit form:
\begin{align}\label{Building_block_polynomials}
    &P_1(h,v,w| r,R) = (h-r^2)(v-r^2),\ \ \ \ P_2(h,v,w| r,R) = (h+v-2r^2), \\\nonumber &P_3(h,v,w| r,R) = (hv-w^2),\ \ \ \ P_4(h,v,w| r,R)=  (h + v - 2w - r^2), \\\nonumber
    &P_5(h,v,w| r,R) = (R^2 - h)(R^2 - v), \ \ \ \ P_6(h,v,w| r,R) = 2R^2 - h - v, \\\nonumber
    &P_7(h,v,w| r,R) = 1.
\end{align}

In particular, for a pair of functions $f_1,f_2$ we consider a sentence $\bm{s}^{(f_1,f_2)} = \bm{m}_1\langle\text{ES}\rangle \bm{m}_2\langle\text{ES}\rangle\ldots\bm{m}_{\ell}\langle\text{EOS}\rangle$, with each monomial term $\bm{m}_i$ being a product of polynomials from \textbf{Equation}~  \ref{Building_block_polynomials}:
\begin{equation*}
 \bm{m}_i = \langle P_1\rangle^{\alpha_{i,1}}\langle \ast\rangle \langle P_2\rangle^{\alpha_{i,2}}\langle \ast\rangle\ldots p^{\alpha_{i,7}}_7.   
\end{equation*}

For any fixed $k\in {1,\ldots,7}$, a non-negative integer $\alpha_{i,k}$ implies the number of repetitions of the polynomial $p_k(h,v,w|r,R)$ in the monomial term $\bm{m}_i$. Notice, as a product of polynomials symmetric with respect to the first and second arguments,  each monomial $\bm{m}_i$ is itself a polynomial symmetric with respect to the first and second arguments and its degree is given as $\text{degree}\left[\bm{m}_i\right] = \sum_{k=1}^7\alpha_{i,k}\text{degree}\left[p_{k}\right]$. 

To explore the search space of all possible sentences in a more structured way, we introduce a parameter $d$ and consider sentences consisting of symmetric monomials with respect to the first two arguments and have a degree at most $d$. Such monomials can be represented as a linear combination in the basis: 
$$\mathcal{B}_d = \{w^{a}(hv)^{b}(h + v)^c: \ \ \ a,b,c\ge 0, \ a + 2b + c \le d\}.$$

\paragraph{SDP Construction: }Given a  sentence $\bm{s}^{(f_1,f_2)}$ consisting of $\ell$ monomial terms $\bm{m}_1, \ldots, \bm{m}_{\ell}$ and a parameter $d$, the player constructs the associated semidefinite program $\text{SDP}(r,R,f_1,f_2)$ in two steps. First, the constraints of $\text{SDP}(r,R,f_1,f_2)$ are expressed in the following form:
\begin{align}\label{Constr_equation}
    &\text{Tr}(\bm{X}^{\mathsf{T}}_1\bm{A}_1(h,v,w|d)) + \text{Tr}(\bm{X}^{\mathsf{T}}_2\bm{A}_2(h,v,w|d)) + \ldots + \text{Tr}(\bm{X}^{\mathsf{T}}_{\ell}\bm{A}_{\ell}(h,v,w|d))= 0,
\end{align}
where each monomial term $\bm{m}_i$ in the sentence introduces a corresponding decision variable $\bm{X}_i$, which is a positive semidefinite matrix. The matrix $\bm{A}_i(h,v,w|d)$ is a polynomial matrix constructed as the product of the monomial $\bm{m}_i$ and a matrix $V_i(h,v,w)$, whose entries consist of pairwise products of polynomials of the basis $\mathcal{B}_{d - \text{degree}[\bm{m}_i]}$. 

The polynomial constraints in \textbf{Equation} \ref{Constr_equation} are converted into standard numerical matrix constraints by substituting symbolic variables $(u,v,t)$ with a set of carefully selected pivot points as noted in~\cite{cohn2022three}. Given those pivot points $(h_j,v_j,w_j)^K_{j=1}$, we can write:
\begin{align*}
    &\text{Tr}(\bm{X}^{\mathsf{T}}_1\bm{A}_{1,j}) + \text{Tr}(\bm{X}^{\mathsf{T}}_2\bm{A}_{2,j}) + \ldots + \text{Tr}(\bm{X}^{\mathsf{T}}_{\ell}\bm{A}_{\ell, j}) = 0, \ \ \ \text{for} \ j\in \{1,\ldots,K\}.
\end{align*}
where $\boldsymbol{A}_{i,j} = \boldsymbol{A}_i(h_j,v_j,w_j|d)$ denotes the resulting numerical coefficient matrix. 

In the second step, we formulate the objective function of $\text{SDP}(r,R,f_1,f_2)$ following the three-point bound estimate $\Delta_{\text{upper}}(r,R, f_1,f_2) \propto \left(f_1(\bm{0}) + f_2(\bm{0}, \bm{0})\right)r^{\text{n}}$. In particular, for a given string $\bm{s}^{f_1,f_2}$ and fixed parameter $r$,  one can construct numerical matrices $\{\bm{C}_{i}\}^{\ell}_{i=1}$ such that $f_1(\bm{0}) + f_2(\bm{0}, \bm{0}) = \sum_{i =1}^{\ell}\text{Tr}(\bm{X}^{\mathsf{T}}_i\bm{C}_i)$. Finally, combining this objective with the constraint expressions described above yields the final semidefinite program associated with parameters $r,R$ and admissible functions $f_1$ and $f_2$:
\begin{align*}\label{SDP-instance}
    &\hspace{4cm}  \ \ \ \min_{\bm{X}_1,\ldots,\bm{X}_{\ell}} \sum_{i=1}^{\ell}\text{Tr}(\bm{X}^{\mathsf{T}}_i\bm{C}_i)\\\nonumber
    &\text{SDP}(r,R,f_1,f_2) \ \ \ \ = \ \ \ \ \ \text{subject to:}\\\nonumber
    &\hspace{4cm} \ \ \ \text{Tr}(\bm{X}^{\mathsf{T}}_1\bm{A}_{1,j}) + \text{Tr}(\bm{X}^{\mathsf{T}}_2\bm{A}_{2,j}) + \ldots + \text{Tr}(\bm{X}^{\mathsf{T}}_{\ell}\bm{A}_{\ell, j}) = 0, \ \ \ \text{for} \ j\in \{1,\ldots,K\}.\\\nonumber
    &\hspace{4cm} \ \ \ \bm{X}_1\succeq 0, \ldots, \bm{X}_{\ell}\succeq 0.
\end{align*}

\subsection*{Bayesian Optimisation}
Bayesian Optimisation (BO) provides a principled framework for optimising expensive black-box functions under tight evaluation budgets. In our setting, each evaluation corresponds to running the model-based search procedure to assess the quality of a candidate certificate ( $r,R$ and/or associated constructed polynomial). Because these evaluations are computationally intensive and the underlying objective landscape is highly non-convex, BO serves as an efficient mechanism for exploration.

At the core of BO is a probabilistic surrogate model; here, a Gaussian Process (GP) that maintains a posterior distribution over the unknown objective function \cite{williams2006gaussian}. Given a set of previously evaluated candidates, the GP provides both a mean prediction $\mu(r,R)$ and an uncertainty estimate $\sigma(r,R)$ for any new point. This uncertainty quantification enables BO to balance exploration (sampling uncertain regions that may contain better solutions) and exploitation (refining regions already known to yield high-quality bounds). To accommodate the heteroskedastic and non-stationary noise that arises in our evaluations, we employ both input and output warping as implemented in the HEBO framework \cite{CowenRivers2022HEBO}.

To select new candidates, we use an acquisition function $a(r,R)$, which maps the GP posterior to a scalar score optimised at each iteration. In our experiments, we employed HEBO, which uses a multi-objective acquisition. At each iteration, BO solves:
\begin{equation*}
    (r_{t+1}, R_{t+1}) = \arg\max_{(r,R)} a_{\text{HEBO}}(r,R;\mu_t(r,R), \sigma_t(r,R)).
\end{equation*}
Given $(r_{t+1}, R_{t+1})$, BO then evaluates the objective at this new point, and updates the GP posterior accordingly.
This process continues until convergence or until the evaluation budget is exhausted.

Within our framework, BO plays two roles: (1) it accelerates the discovery of high-performing candidate $(r,R)$ variables by guiding the search toward promising regions of the parameter space, and (2) it helps explore new constructions of those parameters beyond what mathematicians have considered; see Figure \ref{fig:hebo_r_optim_expensive}. 

\subsection*{Monte-Carlo Tree Search} 
After identifying a promising interval 
$(r,R)$ using BO, we employ Monte Carlo Tree Search (MCTS) to search over the space of admissible polynomial structures that define the semidefinite program (SDP). The goal of this stage is to identify $(f_1,f_2)$ that yield the strongest certified sphere-packing bound within the chosen interval. 

MCTS incrementally builds a search tree in which each node corresponds to a partially specified polynomial (or equivalently, a partially constructed certificate), and each edge corresponds to adding a new token from our vocabulary, respecting the grammar. From the current root state, the algorithm iteratively performs four phases: selection, expansion, simulation, and backpropagation.

During selection, MCTS traverses the tree from the root to a leaf using an Upper Confidence Bound (UCB) criterion that balances exploitation of high-scoring branches with exploration of under-visited ones \cite{kocsis2006bandit}. This allows the algorithm to prioritise polynomial families that have historically yielded strong bounds, while still probing new structures that may outperform existing choices.

In the expansion step, the algorithm adds one or more child nodes by extending the current partial polynomial with additional structural components. These expansions correspond to refining the search space in a controlled and interpretable manner.

The simulation phase evaluates the quality of a completed polynomial by solving the corresponding SDP to obtain its certified sphere-packing bound. Because these evaluations are expensive, we employ lightweight heuristics to prune unpromising branches and use previously computed SDP results to warm-start related simulations, significantly improving efficiency; e.g., by training in lower dimensions and using the solution as a starting node for MCTS.

Finally, in the backpropagation step, the obtained bound is propagated up the tree, updating the value estimates and visitation counts of all ancestor nodes. This feedback mechanism allows the tree to rapidly adapt, amplifying the influence of polynomial families that may yield stronger bounds.

\printbibliography

\end{document}